\documentclass[10pt,letterpaper]{article}

\usepackage{cogsci}
\usepackage{pslatex}
\usepackage{apacite}
\usepackage{graphicx} 
\cogscifinalcopy

\usepackage{kotex}

\title{Are Expressions for Music Emotions the Same Across Cultures?}
 
\author{
    \large
    \textbf{Elif Celen\textsuperscript{1,*}, Pol van Rijn\textsuperscript{1,*}, Harin Lee\textsuperscript{1,2,†}, and Nori Jacoby\textsuperscript{1,3,†}} \\ 
    \textsuperscript{1}Max Planck Institute for Empirical Aesthetics, Frankfurt am Main, Germany  \\
    \textsuperscript{2}Max Planck Institute for Human Cognitive and Brain Sciences, Leipzig, Germany \\
    \textsuperscript{3}Cornell University, New York, United States \\
    \texttt{\{elif.celen, pol.van-rijn, harin.lee, nori.jacoby\}@ae.mpg.de}\\ \\
    \textsuperscript{*} These authors contributed equally to this work \\
    \textsuperscript{†} These authors jointly supervised this work
}

\begin{document}

\maketitle

\begin{abstract}
Music evokes profound emotions, yet the universality of emotional descriptors across languages remains debated. A key challenge in cross-cultural research on music emotion is biased stimulus selection and manual curation of taxonomies, predominantly relying on Western music and languages. To address this, we propose a balanced experimental design with nine online experiments in Brazil, the US, and South Korea, involving N=672 participants. First, we sample a balanced set of popular music from these countries. Using an open-ended tagging pipeline, we then gather emotion terms to create culture-specific taxonomies. Finally, using these bottom-up taxonomies, participants rate emotions of each song. This allows us to map emotional similarities within and across cultures. Results show consistency in high arousal, high valence emotions but greater variability in others. Notably, machine translations were often inadequate to capture music-specific meanings. These findings together highlight the need for a domain-sensitive, open-ended, bottom-up emotion elicitation approach to reduce cultural biases in emotion research.

\end{abstract}

\section{Introduction}

Music uniquely expresses emotions, evokes feelings, and influences our mood, making it a fundamental aspect of human experience \cite{juslin2010, north2008, juslin2008, Park2019-bo}. A central debate in emotion research is whether emotions are innate \cite{darwin1872, ekman1971} or shaped by language and culture \cite{russell2003, barrett2019}. Universalist theories, such as basic emotion theory \cite{ekman1992}, argue that emotions are biologically hardwired and universally recognized, emphasizing a core set of emotions like “happiness,” “sadness,” and “anger.” Supporting this view, studies suggest that basic emotions can be recognized across cultures, even in unfamiliar musical traditions \cite{balkwill1999, fritz2009}. However, other research highlights significant cross-cultural differences in emotional experience \cite{jackson2019}. 

A related question in emotion research is the terminology used to describe emotions. Traditionally, emotion research has focused on a small number of basic emotion categories~\cite{descartes1989passions, ekman1992}. Additionally, emotions have been proposed to exist along dimensions such as arousal (emotional activation) and valence (positive to negative; \citeNP{russell1980}). However, recent studies suggest that describing musical emotions requires a richer vocabulary beyond a limited set of terms and is potentially described by a high dimensional space ~\cite{cowen2019, cowen2021, eerola2013, eerola2025}.

Cross-cultural research on music and emotion has traditionally relied on a small number of terms, often adapted from Western theories \cite{balkwill1999, fritz2009}. However, large-scale studies using online data reveal substantial variability in emotional terms across languages worldwide~\cite{jackson2019}, suggesting that the taxonomy of musical emotions may also differ across cultures. Furthermore, mood perception studies show high agreement on basic attributes like “sad” and “cheerful” across cultures, while terms such as “dreamy” or “love” can have rather culture-specific interpretations~\cite{lee2021}.

How can we characterize music emotion terms across cultures? Previous research has often relied on a limited and biased selection of musical stimuli \cite{fritz2009,balkwill1999} and an arbitrarily chosen set of emotion terms \cite{juslin2008, juslin2010}. Additionally, these terms have primarily been derived from English and other Western languages, while the musical materials studied have predominantly come from Western traditions. These limitations greatly hinder our ability to capture the variability of cultural nuances and discuss its universalities \cite{blasi2022over}.

\begin{figure*}[ht!]
\centering
\includegraphics[width=\textwidth]{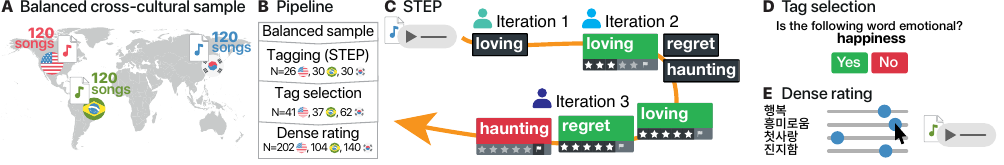}
\caption{Experimental Setup. 
\textbf{A}: Music collection in three cultures. \textbf{B}: Human-in-the-loop pipeline conducted in the following order: (\textbf{C}) STEP paradigm to obtain a list of words per culture, (\textbf{D}) discrete emotionality rating to select emotional words, and (\textbf{E}) dense rating of each stimulus along the select terms.
}
\label{fig:schematics}
\end{figure*}

To address these limitations, we propose a novel pipeline designed to reduce bias and improve cross-cultural comparisons. To mitigate stimulus bias, we carefully sampled a balanced set of 270 songs from Brazil, South Korea, and the US (\citeNP{lee2021}; Figure \ref{fig:schematics}A). We then conducted a series of experiments, recruiting participants from these three countries (Figure \ref{fig:schematics}B) who performed tasks in their native language. Crucially, all participants engaged with the same shared music database, including both music from their own country and from the other two. This design avides an a priori dominance of one culture in the stimulus set.

In each country, a group of participants generated emotion-related tags through an open-ended process (Figure \ref{fig:schematics}C). This allowed participants to freely express relevant emotional terms without being constrained by a predefined taxonomy. The iterative nature of the process also enabled participants to refine, rate, and correct each other’s tags, improving the quality of the final term set. By including diverse music from all three cultures, we ensured that a broad emotional spectrum was covered, avoiding biases that could arise from presenting only a narrow range of stimuli (e.g., if only happy songs were presented, terms like “sad” would not emerge).

A separate group of participants then validated whether the collected terms were truly emotion-related (Figure \ref{fig:schematics}D). This step was necessary because participants in the first phase sometimes reported descriptive musical attributes (e.g., “loud”) alongside emotional terms. By establishing a consensus, we differentiated emotion-related descriptors from general musical characteristics.
	
Since emotion terms may have weak or variable associations with specific stimuli \cite{toivonen2012networks, duran2017, siegel2018, vanrijn2023b}, we accounted for the possibility that multiple terms could apply to a given song (``mixed emotions''). To investigate this, we recruited three additional participant groups—one per culture—who rated the emotions of a subset of songs using the validated taxonomy (Figure \ref{fig:schematics}E). Because all participants listened to the same physical musical stimuli, we could directly compare how emotions were represented across languages.

By analyzing correlations between terms within and across languages, we identified patterns of emotional similarity. Low correlations across songs indicate that participants use these terms in distinct musical contexts, whereas high correlations suggest a stronger semantic connection. We can then use this correlation to map the relationships between emotional terms both within and across cultures.

This pipeline provides a  data-driven approach to understanding music emotions cross-culturally, offering new insights into how emotional expressions in music vary across languages and cultures.

\section{Methods}
\subsection{Song Selection}
We use a dataset of 360 pop songs originally introduced in \cite{lee2021}, which evenly samples songs from Brazil, Korea, and the United States. 
These songs were retrieved from major music charts ensuring a representative selection of popular music trends.

From those 360 songs, 120 songs were presented in each country during the term elicitation phase, consisting of 60 songs from their own culture and 30 songs each from the other two cultures (see `Open-ended tagging using STEP'). Each song was presented for 15 seconds in length.

During the rating phase (see `Dense rating'), a smaller subset of 60 songs (20 songs each from the three countries) were presented uniformly across all three countries. To maximize the diversity of these songs, they were systematically sampled from the entire pool considering the acoustic features of these songs obtained from Spotify (``danceability'', ``energy'', ``loudness'', ``speechiness'', ``acousticness'', ``instrumentalness'', ``liveness'', and ``tempo''), and temporal distribution (two songs per release year for each country).

\begin{figure*}[ht]
\centering
\includegraphics[width=\textwidth]{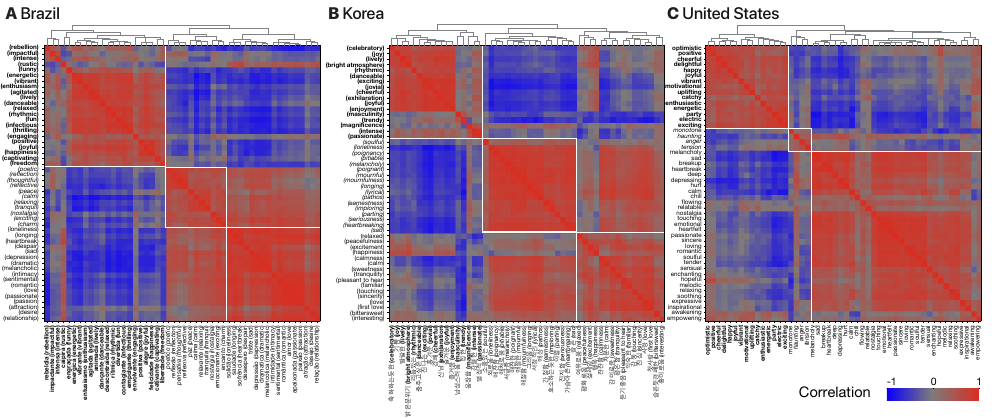}
\caption{Pearson correlation across 50 emotional terms in Portuguese (Brazil, \textbf{A}), Korean (Korea, \textbf{B}), and English (US \textbf{C}). Correlation matrices are sorted by the order in the dendrogram obtained via agglomerative clustering.}
\label{fig:correlation-within}
\end{figure*}

\subsection{Participants}
We recruited independent sets of participants for each experiment we ran. Recruitment criteria included being at least 18 years old, residing in the target country, and speaking the target language as their primary language. Brazilian and US participants were recruited via Prolific (\url{https://prolific.com}), while Korean participants were exclusively recruited through CINT (\url{https://www.cint.com}) due to limited availability on Prolific. Compensation was standardized at £9 per hour for Prolific participants and adjusted to the local minimum wage for CINT participants.

Participants were required to pass a headphone screening task~\cite{milne2021} before participating in experiments that involved listening to music clips.

In addition, to control for language proficiency, all participants completed a vocabulary test~\cite{vanrijn2023}. 

Overall we recruited a total of N=672 participants across all experiments.

\subsection{Procedure}

All experiments were conducted using PsyNet\footnote{Psynet: \url{https://psynetdev.gitlab.io/PsyNet/}} \cite{harrison2020}, an automated recruitment framework for large-scale online studies.
Experimental texts were translated using machine translation and then verified by native speakers from each language. Participants provided informed consent under an approved ethics protocol (Max Planck Ethics Council \#202142).

\begin{figure*}[ht]
\centering
\includegraphics[width=\textwidth]{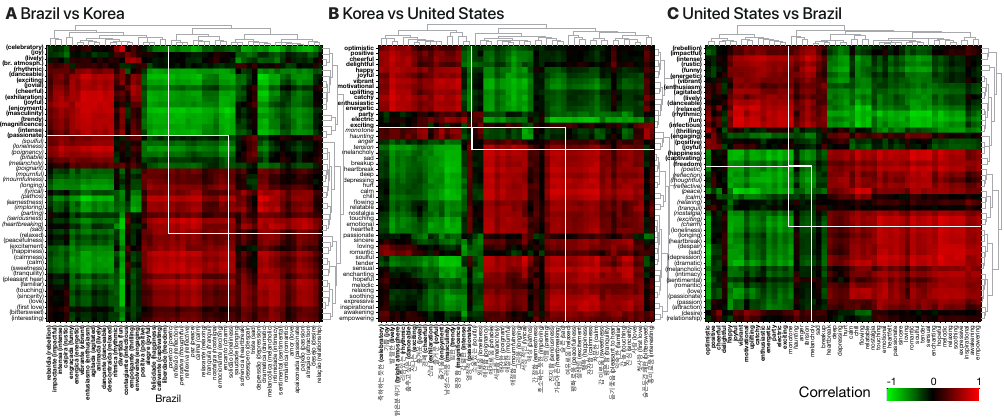}
\caption{Pearson correlation across all three pairs of cultures: (\textbf{A}) Brazil and Korea, (\textbf{B}) Korea and US, and (\textbf{C}) US and Brazil. Correlation matrices are sorted by the order in the dendrogram obtained via agglomerative clustering.}
\label{fig:correlation-across}
\end{figure*}

\subsubsection{Open-ended tagging using STEP.} 
To capture musical experiences without relying on pre-defined taxonomies, we used STEP~\cite{marjieh2023, vanrijn2024}, an open-ended tagging pipeline, where participants generate, evaluate, and refine emotional descriptors over successive iterations (Figure \ref{fig:schematics}C).

In each iteration, (1) participants listened to the full music clip, provided new tags describing their emotional experience, (2) rated existing tags (if previous participants provided tags), and (3) flagged inappropriate ones (e.g., typos, irrelevant responses). They were instructed to use single-word tags that reflect both musical and emotional characteristics, avoiding genre labels and direct transcription of lyrics. Tags were written in their native language, with a 15-character limit, and multiple entries were allowed. Each participant annotated 15 music clips to balance the workload and maintain diversity.

We recruited three groups of participants (US=26, Brazil=30, Korea=30) to annotate 120 randomly selected songs. For each song, after five iterations, this process produced a weighted bag-of-words representation for each song. By analyzing tag co-occurrences, we derived an emergent taxonomy of emotions. This approach is particularly effective for cross-cultural studies, as it iteratively builds culturally grounded semantic representations.

\subsubsection{Tag selection.}
STEP process in each country resulted in 158, 194, and 299 labels for the US, Brazil, and Korea respectively. To create culturally relevant emotional descriptors, we followed a two-step process: cleaning and validation.

In the first step, we cleaned the tags using an automated pipeline (specifically: lemmatizing words, and correcting typos with fuzzy matching, see \citeauthor{niedermann2024} 2024 for more details). Tags containing invalid characters or formats were excluded. After automated cleaning, words with minor spelling variations were manually merged into identical tags. This process resulted in a final list of 150, 186, and 259 emotional descriptors in the US, Brazil, and Korea, respectively (Figure \ref{fig:schematics}D).

To subselect tags referring explicitly to emotional experiences, we conducted a tag selection experiment (US=41, Brazil=37, Korea=62 participants). Each tag was shown to the participants and asked whether the given tag could be used to describe emotions in music or not. An average of 12 emotionality ratings were collected for tags.

Based on this data, we computed the mean emotionality per tag.
Labels that met the 50\% agreement threshold were considered valid emotional terms within that culture.
We subselected the top 50 most emotional terms in each culture.
The final set of 50 terms per language and culture served as the emotional taxonomy for the next step. This number was chosen based on prior studies suggesting that around 50 emotions effectively capture a significant portion of emotional variability~\cite{cowen2019}.

\subsubsection{Dense rating.}
While participants in STEP provide ratings for the relevance of each of the tags in the stimuli, this rating is fairly sparse (i.e., each stimulus receives only a small number of the most appropriate tags). Since emotions can capture more nuanced aspects of music, it is important to account for weakly relevant annotations to better reflect the complexity of emotional responses.

Thus, for a subset of songs (see `Song Selection'), we conducted a dense rating experiment (US=202, Brazil=104, Korea=140 participants) to study the associations across emotion terms for music within and across cultures, (Figure \ref{fig:schematics}E).

For each song, participants randomly received five descriptors from the 50 tags selected in the previous section and provided ratings about its relevance on a 5-point scale. On average, each song and tag was rated 17 times. The order of tags and the presentation of clips were randomized to minimize order effects and potential biases.

\begin{figure*}[ht]
\centering
\includegraphics[width=1.1\textwidth]{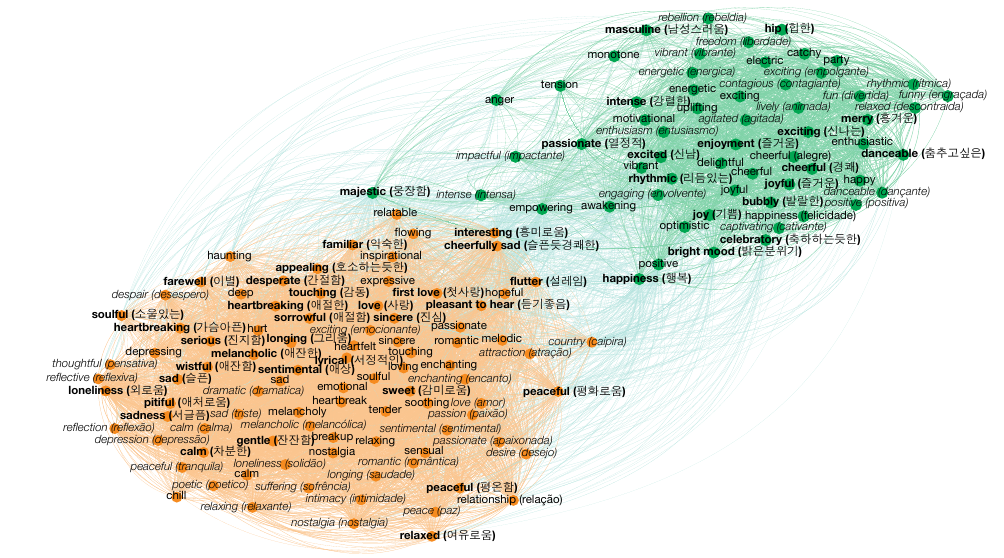}
\caption{Network analysis on the full correlation matrix. Negative correlations are removed. Korean terms are in bold, Portuguese in italics, and English in plain font. The coloring is based on the modularity.}
\label{fig:correlation-network}
\end{figure*}

\section{Results}
\subsection{Within-culture term correlations}
Figure \ref{fig:correlation-within} shows the Pearson correlation matrices of all pairwise comparisons of 50 tags on their ratings within each of the three countries. The results indicate that two major clusters can be seen in all countries, and that the large cluster can be divided into two subclusters.

The first cluster (top left square in each heatmap, Figure \ref{fig:correlation-within}) reflects both high arousal and high valence emotional expressions. Particularly in Brazil this is most prominent, including terms like ``dançante (danceable)'', ``animada (lively)'', and ``vibrante (vibrant)''. Accordingly, terms such as ``춤추고싶은 (danceable)'', ``경쾌 (cheerful)'', and ``신나는 (exciting)'' are found in Korean, 
 while``happy'', ``joyful'', and ``vibrant'' are included in the US. Such cross-cultural consistencies highlight shared taxonomies for expressing energetic and positive emotions when listening to music.

The second cluster (bottom right square in each heatmap) generally reflects low arousal expressions, at the same time, they are more cross-culturally varied and further divided into subclusters. In Brazil, these subclusters reflect high and low valence divide, with high valence subcluster including emotions such as ``calma (calm)'', ``relaxante (relaxing),'' and ``tranquila (tranquil)'', while low valence subcluster includes ``triste (sad)'', ``depressão (depression)'', and ``dramatica (dramatic)''.

In Korea, this subcluster division is even more apparent. One reflects a low valence grouping, described with terms like ``외로움 (loneliness)'', ``서글픔 (melancholy)'', and ``애처로움 (pitiable)'', while the other reflects high valence emotions including ``차분한 (calm)'', ``감미로움 (sweetness)'', and ``평온함 (tranquility)''.

In the US, the subcluster division is more ambiguous, with less distinctive grouping, while blending both positive and negative terms. One (located in the middle) with only a few terms such as ``monotone'', ``tension'', and ``anger''. The other (at the bottom right) represents low arousal emotions, ranging from positive like ``hopeful'', ``relaxing'', to negative like  ``depressing'' and ``sad'', ``breakup'', and ``heartbreak''. 

Empirically, the mean correlations of clustered emotional terms within each group were high (Pearson correlation: US = 0.81, Korea = 0.84, Brazil = 0.87), which confirms that our clustering method adequately captures the distinct groupings. Furthermore, the proportion of variance explained by these clusters, as measured by $R^2$, was also substantial (US = 0.65, Korea = 0.71, Brazil = 0.75), reinforcing the robustness of our approach in organizing emotional terms into coherent and meaningful categories across cultures. 

Together this suggests that while there are broad distinctions for well-established emotional dimensions (i.e., valence, arousal), there are lots of nuances between the terms in languages.

\subsection{Between-culture term correlations}
We next move on to compare how the emotion terms correlate between country pairs, as shown in Figure~\ref{fig:correlation-across}.

As expected, certain emotional terms were used in similar ways across cultures (i.e. are highly correlated to each other). For example, ``sad'' in Korean (``슬픈'') and English, as well as ``romantic'' in Portuguese (``romântica'') and English were strongly correlated (\textit{r} = .73 and .82). Similarly, ``energetic'' in the US sample was highly correlated with ``vibrante (vibrant)'' in the Brazilian sample (\textit{r} = .80).
All cultures had concepts for calmness (``차분한/잔잔함'' and ``calma''), melancholy (``서글픔'' and ``melancólica''), and sadness (``슬픈'' and ``triste''). These results demonstrate alignment in emotional concepts for describing music emotions across cultures.

Surprisingly, however, other emotion terms showed notable differences despite having one-to-one mapping according to dictionary translations.
Using ChatGPT 4.0, we translated all Korean and Portuguese terms to English, and then computed the correlations for all terms with a direct translation available.
These were 22 for Korea and 26 for Brazil (out of 50 terms).
The average correlations were low: for Korean (\textit{r} = .61) and Portuguese (\textit{r} = .59).
For instance, ``열정적 (passionate)'' did not correlate with ``passionate'' (bootstrapped split-halves: \textit{r} = .05 [.00, .20], in Korean:  \textit{r} = .49 [.43, .60] and in English: \textit{r} = .60 [.55, .72]).
At times, terms were even negatively correlated with their direct translation. For example, ``emocionante (exciting)'' and ``exciting'' (\textit{r} = -.50 [-.38, -.54], in Brazil: \textit{r} = .72 [.71, .80] and in English: \textit{r} = .59 [.55, .72]).
These results highlight that dictionary translations are at best proxies of the emotion concept (where the correlation is rather high) but can be ambiguous when describing domain-specific concepts such as music emotions.

We also observed a misalignment between the observed clusters and agglomerative clusters (Brazil--Korea: $R^2 = .38$, Korea--US: $R^2 = .55$, US--Brazil: $R^2 = .41$), as indicated by white lines. For example, in the case of the high arousal, high valence cluster (top left), the pattern in Brazil reflects a structured and focused set of terms such as ``lively'', ``energetic'', and ``cheerful''. In contrast, in Korea and the US, patterns include a wider and more diverse range of terms, not only corresponding to high arousal and high valence but also including ``loneliness'', ``monotone'', and ``melancholy'' which extend beyond the category. This suggests that while participants in Brazil associate such experiences with high arousal and high valence emotions, participants in Korea and the US, interpret the same stimuli through a more varied emotional lens, blending emotions from different dimensions.

\subsection{Correlation network}
So far, we have focussed on correlations of emotion terms across pairs of cultures. In Figure~\ref{fig:correlation-network} we consider the correlations between all three countries and visualise this as a network graph, including only positive correlation values as edges and each term as nodes. The network demonstrates two clear clusters as seen in previous results, and the modularity detection algorithm confirmed this (optimal n clusters = 2; modularity score = 0.345). 

The resulting clusters are highly interpretable with low valence terms in one cluster (orange) and high valence terms in the other (green).
This is aligned with the first two agglomerative clusters in all three countries (ARI: Brazil=.92, Korea=.92, US=.63).
As seen with country-pair correlations in the previous sections, terms that are similar in different languages (e.g., ``peaceful (평온함)'' and ``peace (paz)'') were close together in the network space.

\subsection{In-group effects}
Research in multiple modalities has shown that people agree more with each other when emotional stimulus comes from their own cultural group \cite{elfenbein2002, elfenbein2007, laukka2020, hu2012, hu2014, lee2021}.
To assess if this phenomenon (so-called ``in-group effect'') also exists in our data, we grouped songs based on their origin and average across emotions within each country.

Aligning with previous findings, we found that raters consistently show higher agreement when evaluating songs from their own cultural origin compared to foreign music.
Specifically, Brazilian raters exhibited the highest within-country correlation when rating Brazilian songs, while their agreement was lower for Korean and American songs ($F(2,321) = 20.7, p < .001$, ges = .114). 
Similarly, Korean raters showed significantly greater consistency in rating Korean songs than in rating Brazilian or American songs ($F(2,321) = 20.7, p < .001$, ges = .114).
A comparable pattern was observed for American raters, who demonstrated the strongest agreement when evaluating songs from the U.S ($F(2,321) = 20.7, p < .001$, ges = .114).
These results reinforce the idea that cultural environment and familiarity plays a key role in shaping how listeners perceive musical emotions.

\section{Discussion}

We adopted an open-ended approach to develop taxonomies of emotion for a balanced set of popular music from three different countries. Using these taxonomies, we rated the same database of songs to map similarities and divergences in emotional terms across languages. In all three countries, we identified two broad clusters (Figure~\ref{fig:correlation-within}). The most prominent cluster, present across all cultures, encompassed terms associated with high valence and high arousal. Additionally, certain emotional concepts, such as ``calmness'', ``melancholy'', and ``sadness'', were consistently present across all three cultures. However, secondary clusters varied between cultures. Notably, the clusters did not fully align across cultures (Figure~\ref{fig:correlation-across}).

Our findings indicate that direct machine translations between languages do not always correspond to high correlations in term ratings across musical stimuli. Some direct translations showed no correlation or even negative correlation, suggesting that approaches that do not specifically consider music may fail to fully capture the semantics of emotion across cultures. This underscores the limitations of relying solely on dictionary translations. A predefined taxonomy developed within one culture, and directly translated into another, may not adequately represent the nuanced meanings and contexts of emotional expressions \cite{cowen2019b, cowen2020, cowen2024}.

Our approach offers a significant advantage by allowing for the mapping of emotional terms across languages based on human ratings rather than predefined textual categories. Unlike previous textual methods e.g \cite{jackson2019,thompson2020}, our study incorporates human responses to real, natural music, enabling the identification of patterns specific to music. 

Finally, an in-group effect emerged, with individuals showing stronger agreement when rating music from their own culture.
The significance of this work lies in its ability to produce conceptual results similar to data-driven approaches \cite{cowen2019,cowen2019b,cowen2020}, while addressing biases that have limited their applicability in cross-cultural settings. Although we focused on three cultures, this methodology can be extended to explore a broader range of languages and cultural contexts.

\subsection{Limitations}
One limitation of our study is that we assessed perceived emotions rather than felt emotions \cite{gabrielsson2001emotion}. This distinction is important because we lack direct access to participants’ subjective emotional experiences while listening to music. While felt emotions are certainly of interest, previous research \cite{krumhansl1997exploratory} has demonstrated that felt and perceived emotions in music are correlated. Future studies could validate this relationship using physiological measures such as electrophysiological recordings.
Another limitation is that we only examined three countries. While our approach can be easily extended to additional cultures, our findings may not fully capture global variability. Additionally, our participant pool was drawn from online populations, which may not represent individuals with less exposure to globalized media \cite{jacoby2024commonality}. Although rating experiments can be conducted cross-culturally \cite{athanasopoulos2021harmonic, mcpherson2020perceptual, mcdermott2016indifference}, the iterative nature of our approach, in particular, the STEP stage may require adaptation for field settings.
Furthermore, we focused on popular music, which may have different characteristics from traditional or folk music \cite{mehr2019universality}. However, given that popular music is widely consumed across the globe, our findings remain relevant to understanding emotional expression in music.
Here, we relied on automated and manual cleaning, some of them may be removed when running the STEP process on larger datasets of music.
The rating taxonomy included 50 words, which may have limited cross-language mapping. While this exceeds the empirically determined effective number as suggested by some studies \cite{cowen2020}, expanding the list would require additional participants. However, with large participant pools available across countries, the number of terms can be substantially increased.

\subsection{Conclusion}
Our study combines experimental and computational methods to offer a more detailed understanding of emotions in music. This approach has practical applications, particularly in music recommendation systems that use emotional characteristics to tailor playlists to users’ moods and preferences. Additionally, it can be extended to other domains, including video, speech, vocalization, and images, contributing to broader emotion research \cite{marjieh2023,vanrijn2024}.

\bibliographystyle{apacite}

\setlength{\bibleftmargin}{.125in}
\setlength{\bibindent}{-\bibleftmargin}

\bibliography{main}

\end{document}